\documentclass[twocolumn,10pt]{article}

\usepackage[a4paper,tmargin=25.4mm,bmargin=41.3mm,lmargin=18.5mm,rmargin=18.5mm,footnotesep=5mm]{geometry}
\usepackage[round]{natbib}
\usepackage[utf8]{inputenc}
\usepackage[bottom,multiple]{footmisc}
\usepackage[font=small,labelfont=bf]{caption}
\usepackage[most]{tcolorbox}
\usepackage{abstract}
\usepackage{inconsolata}
\usepackage{graphicx}
\usepackage{fancyhdr}
\usepackage{emptypage}
\usepackage{ragged2e}
\usepackage{hyperref}
\usepackage{listings}
\usepackage{xcolor}
\usepackage{amsmath}
\usepackage{amsfonts}
\usepackage{mathtools}
\usepackage{indentfirst}
\usepackage{adjustbox}
\usepackage{multicol}
\usepackage{xurl}
\usepackage{cancel}
\usepackage{etoolbox}
\usepackage{titlesec}
\usepackage{float}
\usepackage{alltt}
\titleformat{\section}
  {\normalfont\large\bfseries}
  {\thesection}{1em}{}
\titleformat{\subsection}
  {\normalfont\fontsize{11}{13.2}\bfseries\selectfont}
  {\thesubsection}{1em}{}
\titleformat{\subsubsection}
  {\normalfont\normalsize\bfseries}
  {\thesubsubsection}{1em}{}
\titleformat{\paragraph}[runin]
  {\normalfont\normalsize\bfseries}
  {\theparagraph}{1em}{}

\titlespacing*{\section}
  {0pt}{1.5em}{1em}
\titlespacing*{\subsection}
  {0pt}{1.25em}{0.75em}
\titlespacing*{\subsubsection}
  {0pt}{1em}{0.5em}
\titlespacing*{\paragraph}
  {0pt}{0.75em}{0.75em}

\AtBeginEnvironment{thebibliography}{%
  
}

\definecolor{darkgray}{HTML}{505050}
\definecolor{mediumgray}{HTML}{A5A5A5}
\definecolor{lightgray}{HTML}{FAFAFA}
\definecolor{darkred}{HTML}{B41432}
\definecolor{mediumred}{HTML}{D2283C}
\definecolor{lightred}{HTML}{F091A0}

\definecolor{lightblue}{HTML}{EEF5FF}
\definecolor{mediumblue}{HTML}{D4E6FF}
\definecolor{lightgreen}{HTML}{F4FEFF}
\definecolor{mediumgreen}{HTML}{DEECEE}

\hypersetup{ 
    colorlinks=true,
    linkcolor=black,
    filecolor=black,
    citecolor=darkred,
    urlcolor=mediumred,
}

\lstdefinelanguage{pseudocode}{
    morecomment=[l]//,
    morestring=[b][]{\"},
    morestring=[b][]{\'},
    morekeywords={
        INPUT, OUTPUT,
        PROCEDURE, FUNCTION,
        return,
        if, then, else, elif, switch, case, default,
        while, for, in, of, do, continue, break,
        not, or, and,d
        true, false,
        pass,
        error
    }
}
\lstdefinestyle{inlinecode}{
    numbers=none,
    breaklines=true,
    basicstyle=\ttfamily\bfseries\color{black},
    keywordstyle=\ttfamily\bfseries\color{black},
    identifierstyle=\ttfamily\bfseries\color{black},
    stringstyle=\ttfamily\bfseries\color{black},
    commentstyle=\ttfamily\bfseries\color{black}
}

\newtcolorbox{llminput}[1][]{
    breakable,
    sharp corners,
    top=0.5em,
    bottom=0.5em,
    left=0.75em,
    right=0.75em,
    toptitle=0.25em,
    bottomtitle=0.25em,
    boxrule=0.5pt,
    width=\linewidth,
    fonttitle=\scriptsize,
    fontupper=\ttfamily\footnotesize,
    coltitle=black,
    colback=lightblue,
    colframe=mediumblue,
    title=\textbf{MODEL INPUT},
    #1
}
\newtcolorbox{llmoutput}[1][]{
    breakable,
    sharp corners,
    top=0.5em,
    bottom=0.5em,
    left=0.75em,
    right=0.75em,
    toptitle=0.25em,
    bottomtitle=0.25em,
    boxrule=0.5pt,
    width=\linewidth,
    fonttitle=\scriptsize,
    fontupper=\ttfamily\footnotesize,
    coltitle=black,
    colback=lightgreen,
    colframe=mediumgreen,
    title=\textbf{MODEL OUTPUT},
    #1
}

\newtcolorbox{attbox}[1][0.7\textwidth]{
    breakable,
    sharp corners,
    width=#1,
    boxrule=0.7pt,
    colframe=lightgray,
    colback=white,
    top=0.5em,
    bottom=0.25em,
    left=0.5em,
    right=0.25em,
}

\newcommand{\codeinline}[1]{\lstinline[style=inlinecode]{#1}}

\newcommand{\figref}[1]{\textbf{\hyperref[#1]{Figure~\ref{#1}}}}
\newcommand{\lstref}[1]{\textbf{\hyperref[#1]{Listing~\ref{#1}}}}
\newcommand{\tabref}[1]{\textbf{\hyperref[#1]{Table~\ref{#1}}}}
\newcommand{\attref}[1]{\textbf{\hyperref[#1]{Attachment~\ref*{#1}}}}

\renewcommand{\cite}[1]{\citep{#1}}
\title{Beyond Pixels: Exploring DOM Downsampling for LLM-Based Web Agents}

\def\institution{Surfly BV}
\def\authorssubline{Schiepanski and Piël, 2025}

\author{Thassilo M. Schiepanski}
\def\authorcontact{thassilo@surfly.com}
\def\authorTWO{Nicholas Piël}
\def\authorcontactTWO{nicholas@surfly.com}

\makeatletter
\renewcommand*{\maketitle}{
    \renewcommand*\footnoterule{}
    \renewcommand{\headrulewidth}{0pt}
    \lhead{}\chead{\footnotesize\textcolor{darkgray}{\@title}}\rhead{}
    \lfoot{\scriptsize\textcolor{gray}{\authorssubline}}\cfoot{}\rfoot{\thepage}

    \onecolumn
    \thispagestyle{plain}
    \pagestyle{fancy}
    \fancypagestyle{plain}{
        \fancyhf{}
        \rfoot{\thepage}
    }

    \begin{adjustbox}{minipage=0.85\textwidth,center}
        \begin{center}
            \baselineskip=23pt
            {\LARGE\textbf{\@title}}\\
        \end{center}
    \end{adjustbox}
    \vspace{1.75em}

    \begin{adjustbox}{minipage=0.7\textwidth,center}
        \begin{center}
            \begin{tabular}{@{} l l l }
                \begin{tabular}{@{} l}
                    {\normalsize\@author\vspace{0.15em}} \\
                    {\normalsize\authorcontact}
                \end{tabular} &
                &
                \begin{tabular}{@{} l}
                	{\normalsize\authorTWO\vspace{0.15em}} \\
                   {\normalsize\authorcontactTWO}
                \end{tabular}
            \end{tabular}
        \end{center}
        \vspace{0.5em}

        \begin{centering}
            {\small\institution\par}
        \end{centering}
        \vspace{1.5em}

        \begin{centering}
            {\url{https://github.com/webfuse-com/D2Snap}\par}
        \end{centering}
        \vspace{-1.5em}

        \renewcommand*\abstractname{}
        \setlength{\absleftindent}{0mm}
        \setlength{\absrightindent}{0mm}
        \begin{abstract}
            \normalsize
            The advent of large language models (LLMs) has sparked an evolution of autonomous web browsing agents: given a web browsing task and serialised user interface (UI) state, an LLM is expected to suggest input actions that incrementally solve the given task. The central challenge lies in serialising UI state for LLMs. Web agents have increasingly relied on grounded GUI snapshots -- screenshots augmented with visual cues -- favoured for their modest input token footprint. DOM snapshots -- serialised as HTML -- represent a compelling alternative that leverages previously demonstrated HTML interpretation capabilities of LLMs. Their excessive input token footprint, however, has precluded reliable deployment with web agents to date.

We propose D2Snap -- an algorithm to downsample the DOM, premised on preserving actionability and actionability-discriminating features. We evaluate D2Snap-downsampled DOM snapshots using a snapshot-variant web agent (GPT-4o) on a dataset sampled from Online-Mind2Web. Whilst 42\% of raw DOM snapshots exceed the model's context window ($128\times 10^3$ tokens), all D2Snap-downsampled DOM snapshots of our reference configuration fit, at a mean context utilisation of 16.5\%. Against the 67\% success rate of a grounded GUI snapshot baseline, our configuration attains 73\% ($+5.8$\%pt; 95\% CI $-13.6$ to $+26.0$\%pt; McNemar, $p = 0.47$), excluding a deficit (one-sided 95\%) beyond 11\%pt. Image input moreover appears to add little to snapshot utility; grounding text alone attains 62\% ($-5.8$\%pt; McNemar, $p = 0.37$).
        \end{abstract}
    \end{adjustbox}
}
\makeatother

\begin{document}
    \pagenumbering{arabic}

    \onecolumn
    \maketitle

    \vspace{0.975em}
    \begin{figure}[!ht]
        \centering
        \includegraphics[width=0.6975\textwidth]{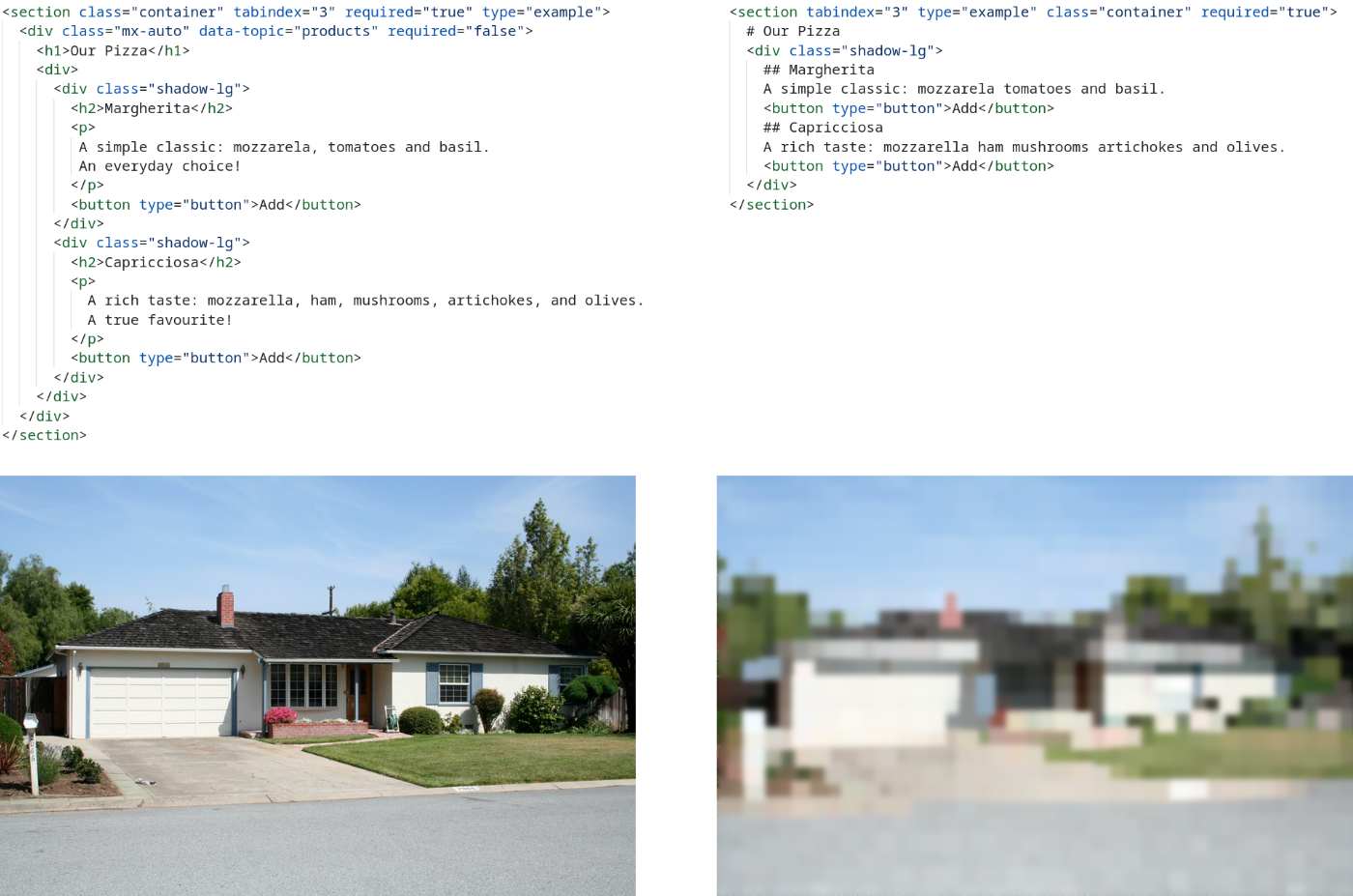}
        \label{fig:0-1}
    \end{figure}

    \twocolumn
    \section{Introduction}

Web-based user interfaces have traditionally been human-centric and, consequently, developed primarily around visual output and manual input. Artificially intelligent agents that autonomously browse the web on behalf of humans (hereafter referred to as web agents) have been a longstanding subject of research~\cite{AmandiArmentano2004, GurFuruta2023, HeYao2024}. Such agents are constrained to interact with human-facing user interfaces, as application programming interfaces are often unavailable, incomplete, or restricted~\cite{Booking}. Conventional AI agents have largely relied on heuristic search over a formal model of the problem domain~\cite{PooleMackworth2017}. Because web application UIs are highly complex and diverse, constructing a model adequate for human-centric web browsing is non-trivial. Large language models have recently enabled a new approach to building web agents. In this paradigm, an LLM serves as a reasoning backend: given serialised UI state (hereafter referred to as a snapshot), it is expected to suggest input for advancing a specified web browsing task~\cite{DengGu2023, OpenAI_0}.

\subsection{Problem}

Web browser automation frameworks expose perception and actuation interfaces to capture output and send input, respectively~\cite{Selenium, Playwright_0}. A central problem in developing web agents (hereafter used synonymously with LLM-based web agents) is context engineering~\cite{MeiYao2025, Anthropic_0}: how best to facilitate an LLM's interpretation of a web page UI to elicit accurate input actions, each specified by a type and a target (e.g., \texttt{click} on a specific \texttt{button}).

GUI (graphical UI) snapshots, i.e., screenshots, emulate what humans see when viewing a web page. Image input to LLMs is downsampled, which decouples input byte size from input token count~\cite{OpenAI_1, OpenAI_2, OpenAI_3}. As a result, GUI snapshots of representative real-world web pages fit comfortably within the context window of many frontier LLMs today~\cite{OpenAI_4, Anthropic_1}. LLM vision capabilities still lack pixel precision as image downsampling does not systematically preserve actionable elements. For that reason, state-of-the-art web agents have increasingly been premised on grounded GUI snapshots -- screenshots enhanced with visual cues that survive downsampling~\cite{BrowserUse_0, BrowserUse_1, YangZhang2023} (see \attref{att:a}). Text-based snapshots -- DOM snapshots in particular -- present a viable alternative to GUI snapshots. The DOM is a web browser's runtime model of a web page, and HTML is the DOM's canonical serialisation format~\cite{WHATWG_1}. LLMs have demonstrated the ability to interpret HTML and even navigate UIs indirectly encoded as HTML~\cite{GurNachum2022}. However, DOM snapshots of representative real-world web pages routinely exceed the context window of many LLMs today~\cite{Webfuse}.

Enabling DOM snapshots for use with web agents therefore motivates pre-processing techniques, analogous to image pre-processing that LLM providers already implement~\cite{OpenAI_1}. Such techniques must reduce a snapshot's size without compromising its utility as input to a web agent's LLM backend. Preserving a valid DOM is desirable, ultimately, to allow further processing, such as translation to an accessibility tree. Naïve truncation of the serialised DOM removes actionable elements indiscriminately and breaks HTML syntax.

\subsection{Contributions}

In this work, we propose \textit{D2Snap}, an algorithm for downsampling the DOM. Intended as a pre-processing technique for use with web agents, it locally consolidates DOM nodes such that the output's utility as a snapshot is largely preserved -- trading fidelity for size. Crucially, it remains a valid DOM.

We evaluate D2Snap through a minimal, snapshot-variant agent (\textit{GPT-4o}~\cite{OpenAI_4, OpenAI_5} backend) on web browsing tasks derived from \textit{Online-Mind2Web}~\cite{XueQi2025}. To avoid subjective judgement, we handcraft a multimodal snapshot dataset along verified solution trajectories (52 snapshots, 18 trajectories), annotated with ground-truth action targets.

Raw DOM snapshots exceed the model's context window ($128 \times 10^3$ tokens) in 42\% of the cases, at a mean model context utilisation of 109\%. All snapshots of our reference configuration -- low retention of elements, high retention of attributes, moderate retention of text -- fit, at a utilisation of 16.5\% against 3.0\% for a grounded GUI snapshot (Set-of-Mark)~\cite{BrowserUse_0, YangZhang2023} baseline. Its 73\% success rate is not significantly different from the baseline's 67\% rate ($+5.8$\%pt, 95\% CI $-13.6$ to $+26.0$\%pt; McNemar, $p = 0.47$), excluding a deficit (one-sided 95\%) beyond 11\%pt. Our results support that HTML conveys salience: flattening the DOM to text with inlined actionable elements significantly underperforms our reference ($-26.9$\%pt, $p = 0.002$). Baseline snapshots without images attain 62\% ($-5.8$\%pt; $p = 0.37$), indicating that image input adds little to snapshot utility.
\section{D2Snap}

Downsampling is a technique to reduce data through local, lossy consolidation, such that encoded features are largely preserved~\cite{GonzalezWoods2018}. A digital image can be locally consolidated by averaging an $n$-tile of pixels -- the depicted object would remain recognisable~\cite{LeCunBottou1998}. We transfer this principle to the DOM: it can analogously be locally consolidated by merging $n$-subtrees of nodes into a single 'average' node. What constitutes a feature thereby depends on the intended purpose of the downsampled DOM; in our case, features are implied by the downsampled DOM's utility as a snapshot for a web agent's LLM backend. A gradual increase of the downsampling ratio should yield a gradual DOM reduction. Since HTML lexemes have unbounded length, however, monotonic size reduction is a strong aim, and linear size reduction a weak, best-effort aim.

\subsection{Algorithm}

We propose D2Snap {\small(\textit{\underline{D}ownsample \underline{D}OM \underline{Snap}shots})} -- an informed algorithm for downsampling the DOM, designed to preserve (A) its utility as a web agent snapshot, and (B) syntactic validity. It figuratively averages redundant nodes of three types: elements, attributes and text~\cite{WHATWG_0}. To control downsampling per type, we introduce three parameters over the unit interval: the ratios $r_e$ for elements and $r_t$ for text, and the discarding threshold $r_a$ for attributes ($r_e, r_a, r_t \in [0, 1]$, 1 $\equiv$ maximal downsampling). \attref{att:b} lists examples of D2Snap-downsampled DOM snapshots.

The DOM is a tree structure~\cite{WHATWG_0} -- size reduction lies in reducing the hierarchy rendered by nested elements. D2Snap averages elements by merging adjacent elements. The parameter $r_e$ governs the removed height fraction to target a DOM height of $\max(1, \lceil h(\mathrm{DOM}) \cdot (1 - r_e) \rceil)$: merge levels are distributed evenly by a Bresenham gate~\cite{Bresenham1965}. A virtual root node is implied, exempt from consolidation, to admit full DOM flattening. A merge describes substituting the deeper candidate element by its own children. This concept of averaging rests on two assumptions: (1) structural segregation decreases with depth, and (2) most attributes are reflexive or forward-transitive (e.g., \texttt{id}, or \texttt{hidden}~\cite{WHATWG_1}, respectively). The converse of (2) does not hold -- an invisible element can have a visible parent. As downsampling trades fidelity for size, we accept that attributes might become fuzzy or be swallowed\footnote{Pruning ineffective (e.g., hidden) subtrees from the DOM beforehand would mitigate attribute semantics drift.}.

\begin{lstlisting}[caption={D2Snap consolidates element nodes by substituting an element with its children, distributed evenly through an $r_e$-ratio Bresenham gate. The root level is exempt, retaining a valid DOM at $r_e = 1$. Absolute measures: actionable elements are preserved as-is, text-formatting elements are translated to Markdown.},
label={lst:2-1},
language=pseudocode,
mathescape]
switch Class(ELEMENT_NODE):
  case Class.ACTIONABLE:
    break
  case Class.TEXT_FORMATTING:
    TEXT_NODE $\leftarrow$ Markdown(ELEMENT_NODE)
    Replace(ELEMENT_NODE, TEXT_NODE)
    break
  default:
    d $\leftarrow$ Depth$_{\mathrm{frozen}}$(ELEMENT_NODE)
    if d > 1 and $\lfloor \text{d} \cdot r_e \rfloor$ > $\lfloor (\text{d} - 1) \cdot r_e \rfloor$:
      Replace(ELEMENT_NODE, ELEMENT$_{\mathrm{Children}}$)
      break
\end{lstlisting}

\paragraph{\mbox{Actionable Elements.}} Elements that semantically afford actionability represent atoms of the DOM-encoded UI -- that is, triggers for UI state transitions. Element consolidation idiomatically transfers the concept of downsampling to the DOM. However, as we design D2Snap for use with web agents, we strictly exclude actionable elements from consolidation. For atomicity, actionable elements, such as buttons, should not inflate with content, particularly from parents. Also, actionable elements presumably hold no deep container substructure, so their exclusion only marginally compromises the targeted DOM height.

\paragraph{\mbox{Text-Formatting Elements.}} As an absolute size reduction measure, we include translation of HTML-serialised text-formatting elements to token-efficient Markdown equivalents~\cite{Gruber2004, GitHub}. Most aggressive HTML syntax downsampling ($r_e = r_a = 1, r_t = 0$) then produces a Markdown representation with inlined actionable elements -- an input-aware alternative to existing Markdown snapshots~\cite{Ango2025}.

\smallskip

\lstref{lst:2-1} is a pseudocode description of D2Snap's element node sub-procedure.

\subsubsection{Attribute Nodes}

Attributes are leaf nodes attached to elements, so reducing the DOM in elements implies reducing attributes. There is no evident analogue of local consolidation for attributes, e.g., regarding centrality. To avoid unconditioned survival of heavy attributes, D2Snap discards attributes whose relevance falls below $r_a$. We are aware of no authoritative such relevance scoring, so we define it as a heuristic represented through an empirical parameter $\mathcal{A}$ -- a lookup table (see \attref{att:c}) or scoring function ($\in [0, 1]$). If $\mathcal{A}$ not only scores attributes by relevance, but also calibrates those scores against global attribute frequency, the threshold $r_a$ functions as a latent downsampling ratio -- ideally governing removal of the $r_a$-fraction of attributes from the DOM. \lstref{lst:2-2} is a pseudocode description of D2Snap's attribute node sub-procedure.

\begin{lstlisting}[caption={D2Snap removes attributes that score below the threshold $r_a$. If $\mathcal{A}$ is calibrated against global attribute frequency, $r_a$ functions as a latent downsampling ratio.},
label={lst:2-2},
language=pseudocode,
mathescape]
if $\mathcal{A}$(ATTRIBUTE_NODE$_{\mathrm{Name}}$) < $r_a$:
  Remove(ATTRIBUTE_NODE)
\end{lstlisting}

\subsubsection{Text Nodes}

Text nodes embed natural language into the DOM. D2Snap consolidates text by reducing its size at the sentence level~\cite{RaiBorah2021}. Of the many text-summarisation techniques available, D2Snap uses the \textit{TextRank} algorithm~\cite{MihalceaTarau2004} for a generalist setup, pruning the floored $r_t$ fraction of sentences with the lowest centrality within the text body, whilst preserving their original order\footnote{TextRank on sentences is meaningless for few sentences. For short texts, it can be substituted, e.g., by truncation.}. Single-sentence text nodes, as common for buttons or links, are hence retained. \lstref{lst:2-3} is a pseudocode description of D2Snap's text node sub-procedure.

\begin{lstlisting}[caption={D2Snap consolidates text nodes on a sentence level: text is tokenised into sentences and ranked by centrality using the TextRank algorithm. The lowest-ranking $r_t$ fraction is discarded; surviving sentences are re-emitted to the text node in their original order. At least one sentence is always retained.},
label={lst:2-3},
language=pseudocode,
mathescape]
SENTENCES $\leftarrow$ Tokenize(TEXT_NODE$_{\mathrm{TextContent}}$)
SENTENCES$_{\mathrm{ranked}}$ $\leftarrow$ TextRank$_{\textrm{Sentence}}$(SENTENCES)
SENTENCES$_{\mathrm{selected}}$
  $\leftarrow$ SENTENCES$_{\mathrm{ranked}}$[0 : $\max(\lceil(1 - r_t) \cdot |\text{SENTENCES}|\rceil, 1)$]
SENTENCES$_{\mathrm{re-ordered}}$ $\leftarrow$ Order$_{\textrm{Natural}}$(SENTENCES$_{\mathrm{selected}}$)
TEXT_NODE$_{\mathrm{TextContent}}$ $\leftarrow$ Join(SENTENCES$_{\mathrm{re-ordered}}$)
\end{lstlisting}

\attref{att:d} displays a full pseudocode description of D2Snap.
\section{Evaluation}

Our experimental setup to evaluate D2Snap isolates the snapshot representation as the independent variable, and its utility as LLM input as the dependent variable. Our principal research questions are:

\begin{itemize}
    \item \textbf{RQ~1.} Do D2Snap-downsampled DOM snapshots fit comfortably within a frontier LLM's context window? We operationalise a comfortable fit as mean model context utilisation (CU$_{\%}$) of at most 20\%; leaving headroom for two-snapshot history ($3\times$ 20\%), additional history context ($2\times$ 10\%) and a system prompt (20\%).
    \item \textbf{RQ~2.} Do D2Snap-downsampled DOM snapshots allow a frontier LLM to suggest input action targets that advance real-world web browsing tasks?
\end{itemize}

\subsection{Evaluation Dataset}

\begin{figure*}[!t]
    \includegraphics[width=\textwidth]{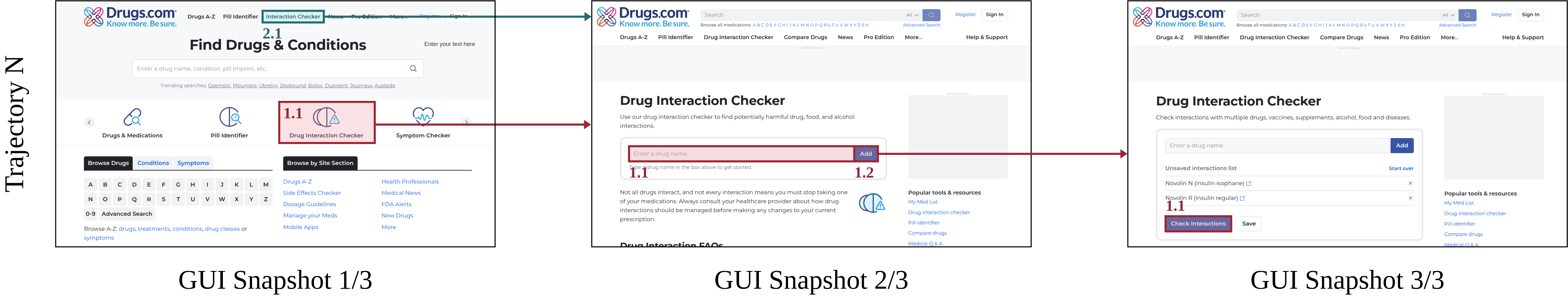}
    \caption{Annotated solution trajectory of UI states encoded as GUI snapshots from our dataset on \textit{drugs.com} for the task: \textit{``Check the interaction between Novolin N and Novolin R''}. Coloured bounding boxes represent annotated action targets. A group of equally coloured boxes forms a coherent target set. A web agent must interact with all elements of a single coherent target set to transition across the UI state trajectory. Our dataset contains full-page GUI snapshots.}
    \label{fig:3-1}
\end{figure*}

To this end, we construct a dedicated snapshot dataset, sourced from Online-Mind2Web~\cite{XueQi2025}. We randomly sample 6 easy, 6 medium, and 6 hard web browsing tasks.

In a first pass, a human (non-author) is asked to solve each sampled task, for which we collect a solution trajectory of snapshots\footnote{Snapshots are taken before each input~\cite{OSUNLP_0}.}. Each snapshot is a triple of independent representations: a DOM snapshot, a GUI snapshot, and a grounded GUI snapshot, emulating the Set-of-Mark~\cite{YangZhang2023} approach of \textit{Browser Use}~\cite{BrowserUse_0}. We verify all trajectories via \textit{WebJudge}~\cite{OSUNLP_2}; all 18 trajectories passed verification. The resulting dataset contains a total of 52 tuples of tasks and snapshot-triples -- each snapshot along a trajectory is treated as an independent sub-task.

In a second pass, two software engineers (non-authors) independently annotate every snapshot triple, identifying all coherent sets of input action targets. A coherent action target set is a minimal set of UI elements that requires input (e.g., clicking or typing) to advance a related web browsing task, i.e., transition between adjacent, encoded UI states. A snapshot may admit multiple coherent target sets (e.g., a redundant link in the \texttt{header} and the \texttt{footer}). An annotation takes the form of bounding-box coordinates for a GUI snapshot, supplemented with a numerical identifier (if available) for a grounded GUI snapshot, and a unique CSS selector for a DOM snapshot. For assessing snapshot utility, we do not verify action target arguments, nor order within coherent action target sets. Annotations ($A$, $B$) yield an inter-annotator agreement of F1 = 0.89\footnote{Pairwise F1 between both annotators' action target sets ($|A| = 471$, $|B| = 499$, $|A \cup B| = 538$, $|A \cap B| = 432$). Two annotated action targets match if: (1) they share the same snapshot, (2) they share the same representation, and (3), depending on the representation, their bounding boxes overlap by $\text{IoU} \ge 0.75$, their numeric IDs are equal, or their CSS selectors resolve to the same element.}. Given high agreement, we take either annotator's targets to be valid and build the evaluation reference from the union\footnote{Two matched bounding boxes are united as the minimal rectangle enclosing both.}, i.e., ground-truth action targets against which LLM-suggested targets can be matched. \figref{fig:3-1} shows an example annotation for a trajectory of GUI snapshots. \attref{att:e} excerpts a dataset reference annotation. 

\subsection{Evaluation Framework}

The evaluation of snapshot representations is designated to deploy a web agent across each tuple from the dataset, once per representation. The web agent LLM backend's response is counted as a success if the suggested action targets form a superset of any coherent target set in the reference\footnote{Elements are compared with slight tolerance: for GUI snapshots, we inflate referenced bounding boxes by 10 pixels to each side; for DOM-based snapshots, we additionally accept a referenced node's parent. The latter accommodates input event bubbling behaviour~\cite{StchurSteiner2021}. Although expressed in different units, both tolerances absorb representation-specific ambiguity: pixel-level annotation jitter (GUI), and handler uncertainty (DOM).}\footnote{The agent could trivially satisfy the success criterion by excessively suggesting action targets. We mitigate such degenerate suggestions by rejecting target sets exceeding $3\times$ the size of the largest coherent reference set. The buffer accounts for additional, non-interfering action targets (e.g., an info-box). Across our experiments, the ceiling was reached at most twice per snapshot representation.}. At the aggregate level, the evaluation success rate is computed over all 52 task-snapshot tuples per representation.

\subsection{Experimental Setup}

\begin{figure*}[!t]
    \centering
    \includegraphics[width=0.95\linewidth]{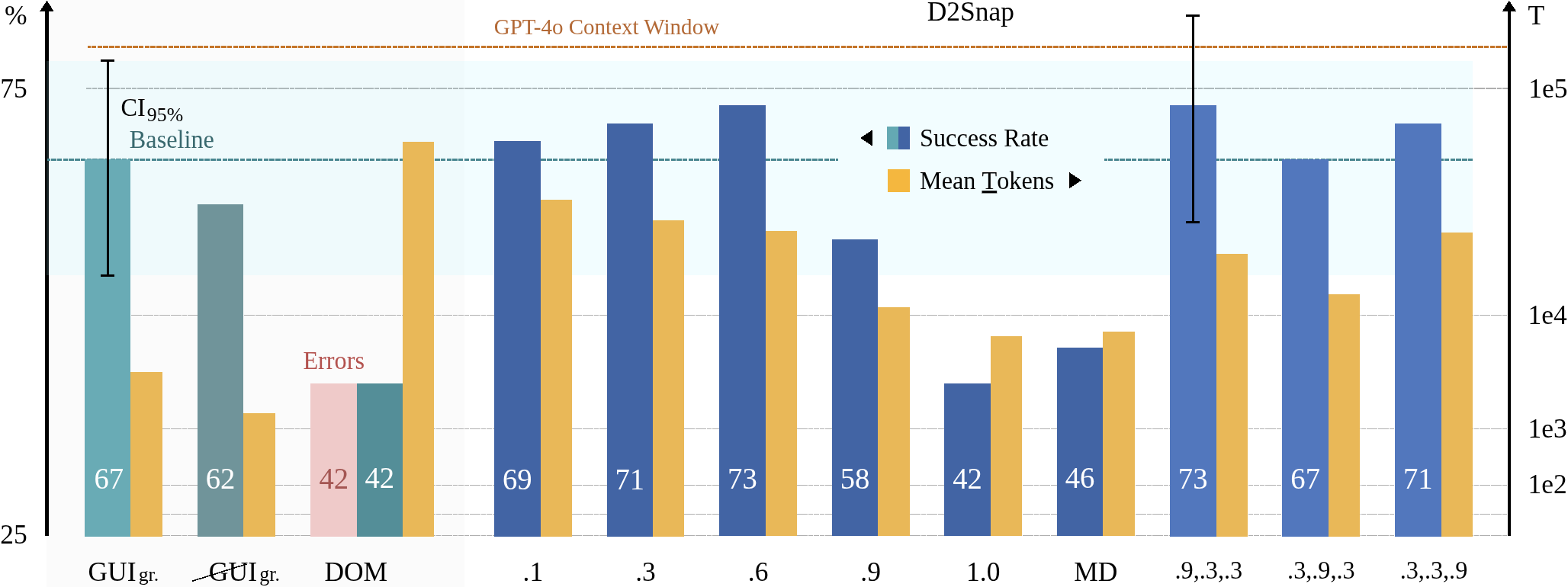}
    \caption{Agent success rates and mean token sizes across snapshot representations plotted as a bar chart. Grounded GUI snapshots (GUI$_{\mathrm{gr.}}$) represent the baseline (67\%, 95\% Wilson CI: 54\% to 78\%). 42\% of raw DOM snapshots exceeded the model's context window (GPT-4o, $128 \times 10^3$ tokens). Every D2Snap-downsampled snapshot fits. Success with D2Snap-based representations is not significantly different from the baseline (e.g., D2Snap$_{.9,.3,.6}$, 73\%; $+5.8$\%pt, 95\% CI $-13.6$ to $+26.0$\%pt; McNemar, $p = 0.47$). Grounding text alone (\cancel{GUI}$_{\mathrm{grounded}}$) reaches 62\% (McNemar, $p = 0.37$).}
    \label{fig:3-2}
\end{figure*}

We deploy a minimal web agent -- merely a model with a snapshot-variant system prompt -- to attribute performance differences to snapshot representations. As the backend, we choose GPT-4o (\texttt{gpt-4o-2024-11-20}~\cite{OpenAI_4}), a multimodal, general-purpose model, with a context window of $128 \times 10^3$ tokens ($\equiv 100\%\ \text{CU}_{\%}$)\footnote{For reproducibility, agent response temperature is fixed at $0$. Results per snapshot representation are averaged across three runs; booleans are averaged by majority.}. Rather than establishing heuristic attribute scoring manually, we elicit it directly from the model, independently of the dataset, which presupposes the model's capacity to interpret HTML semantics -- the enabling assumption of LLM-based DOM interpretation. Additionally, we write a system prompt template instructing the model to suggest all action targets that advance a given task on a web page encoded by a given snapshot, and substitutions specific to each snapshot representation. With DOM snapshots, for instance, element targeting is described by means of CSS selectors. \attref{att:c} records our experimental attribute scoring lookup table, including eliciting model prompts. \attref{att:f} records the exact system prompt templates used in our experiments.

\paragraph{\mbox{Snapshot Representations.}} As the baseline snapshot technique, we adopt grounded GUI snapshots, as implemented by \textit{Browser Use}~\cite{BrowserUse_1} following Set-of-Mark~\cite{YangZhang2023} (A). Our evaluation compares the baseline against its grounding text alone, i.e., without the image (B), raw GUI snapshots (C), and raw DOM snapshots (D). It covers a range of D2Snap-downsampled DOM snapshots compiled from raw DOM snapshots from the dataset (E, F), including a linearised Markdown representation with inlined actionable HTML (G). \attref{att:a} shows an example of a GUI snapshot and its equivalent grounded GUI snapshot. The evaluated snapshot representations are referred to by class notation:

\bgroup
\def\arraystretch{1.5}
\begin{center}
    \begin{tabular}{@{}llp{5.5cm}@{}}
        A. & \textbf{GUI$_{\mathrm{grounded}}$} & \textbf{Baseline} \\
        B. & \textbf{\cancel{GUI}$_{\mathrm{grounded}}$} & GUI$_{\mathrm{grounded}} \setminus \text{image}$ \\
        C. & \textbf{GUI} & raw screenshots \\
        D. & \textbf{DOM}\footnote{We remove redundant whitespace from all DOM-based snapshots.} & raw serialised DOM (HTML) \\
        E. & \textbf{D2Snap$_{r_e, r_a, r_t}$} & $r_e, r_a, r_t \in [0, 1]$\footnote{We remove evidently irrelevant elements (e.g., \texttt{meta}~\cite{WHATWG_1}) from D2Snap-downsampled DOM snapshots.} \\
        F. & \textbf{D2Snap$_r$} & $r = r_e = r_a = r_t$ \\
        G. & \textbf{D2Snap$_{\mathrm{MD}}$} & $r_e = r_a = 1, r_t = 0$ \\
    \end{tabular}
\end{center}
\egroup

\subsection{Experimental Results}

\bgroup
\def\arraystretch{1.2}
\begin{table*}[!t]\footnotesize
    \centering
    \begin{tabular*}{\textwidth}{@{\extracolsep{\fill}}l *{9}{r}}
        \hline
        & \multicolumn{2}{c}{\textbf{Success} \scriptsize (\%)} & \multicolumn{5}{c}{\tiny$\overline{X}$ \footnotesize \textbf{Size} \scriptsize ($\times 10^3$)} & \multicolumn{2}{c}{\tiny$\overline{X}$ \footnotesize \textbf{Latency}} \\
        & Rate & \multicolumn{1}{r |}{Wilson CI$_{95\%}$} & Bytes (B) & Tokens (T) & P$_{5,95}$ (T) & $\text{CU}_{\%}$ & \multicolumn{1}{r |}{\scriptsize$\Delta_{\mathrm{Raw}}$} & \scriptsize$\hat{\Delta}$ & SEM \\
        \hline
        \textbf{GUI$_{\mathrm{grounded}}$} \scriptsize (Baseline) & \textbf{67.3} & \scriptsize [53.8, 78.5] & 2,378.22 & 3.78 & 0.65, 11.38 & \textbf{3.0} & -- & \textbf{3.42} & 0.27 \\
        \textbf{\cancel{GUI}$_{\mathrm{grounded}}$} & \textbf{61.5} & \scriptsize [48.0, 73.5] & 5.84 & 1.46 & 0.36, 3.84 & \textbf{1.1} & -- & 1.00 & 0.07 \\
        \textbf{GUI} & 1.9 & \scriptsize [0.3, 10.1] & 2,349.33 & 2.29 & 0.17, 7.63 & 1.8 & -- & \textbf{3.40} & 0.24 \\
        \hline
        \textbf{DOM} & \textbf{42.3} & \scriptsize [29.9, 55.8] & 559.10 & 139.78 & 13.39, 498.39 & \textbf{109.2} & 0.93 & 3.05 & 0.28 \\
        \hline
        \textbf{D2Snap$_{.1}$} & 69.2 & \scriptsize [55.7, 80.1] & 123.00 & 30.73 & 5.21, 93.61 & 24.0 & 0.28 & 2.31 & 0.22 \\
        \textbf{D2Snap$_{.3}$} & 71.2 & \scriptsize [57.7, 81.7] & 104.38 & 26.07 & 4.69, 84.02 & 20.4 & 0.24 & 2.50 & 0.25 \\
        \textbf{D2Snap$_{.6}$} & \textbf{73.1} & \scriptsize [59.7, 83.2] & 93.37 & 23.32 & 3.93, 76.74 & 18.2 & 0.20 & 2.46 & 0.24 \\
        \textbf{D2Snap$_{.9}$} & 57.7 & \scriptsize [44.2, 70.1] & 42.55 & 10.62 & 1.73, 30.47 & 8.3 & 0.11 & 1.72 & 0.17 \\
        \textbf{D2Snap$_{1.0}$} & \textbf{42.3} & \scriptsize [29.9, 55.8] & 18.77 & 4.69 & 1.11, 10.71 & \textbf{3.7} & \textbf{0.06} & \textbf{1.38} & 0.11 \\
        \hline
        \textbf{D2Snap$_{\mathrm{MD}}$} & \textbf{46.2} & \scriptsize [33.3, 59.5] & 20.96 & 5.24 & 1.13, 11.42 & 4.1 & \textbf{0.06} & 1.49 & 0.15 \\
        \hline
        \textbf{D2Snap$_{.9,.3,.3}$} & \textbf{73.1} & \scriptsize [59.7, 83.2] & 85.83 & 21.46 & 3.71, 67.00 & 16.8 & 0.19 & 2.14 & 0.19 \\
        \textbf{D2Snap$_{.3,.9,.3}$} & 67.3 & \scriptsize [53.8, 78.5] & 48.88 & 12.22 & 1.91, 32.79 & 9.5 & 0.13 & 1.79 & 0.15 \\
        \textbf{D2Snap$_{.3,.3,.9}$} & 71.2 & \scriptsize [57.7, 81.7] & 102.38 & 25.59 & 4.52, 83.57 & 20.0 & 0.23 & 2.18 & 0.19 \\
        \hline
        \textbf{D2Snap$_{.3,.6,.9}$} & 69.2 & \scriptsize [55.7, 80.1] & 100.45 & 25.11 & 4.37, 83.32 & 19.6 & 0.22 & 2.42 & 0.24 \\
        \textbf{D2Snap$_{.3,.9,.6}$} & 65.4 & \scriptsize [51.8, 76.8] & 47.55 & 11.89 & 1.89, 32.24 & 9.3 & 0.12 & 1.76 & 0.16 \\
        \textbf{D2Snap$_{.6,.3,.9}$} & \textbf{73.1} & \scriptsize [59.7, 83.2] & 94.56 & 23.64 & 4.14, 76.77 & 18.5 & 0.21 & 2.08 & 0.20 \\
        \textbf{D2Snap$_{.6,.9,.3}$} & 55.8 & \scriptsize [42.3, 68.4] & 47.03 & 11.76 & 1.83, 32.12 & 9.2 & 0.12 & 1.79 & 0.17 \\
        \textbf{D2Snap$_{.9,.3,.6}$} & \textbf{73.1} & \scriptsize [59.7, 83.2] & 84.50 & 21.13 & 3.69, 66.73 & \textbf{16.5} & 0.19 & \textbf{2.42} & 0.23 \\
        \textbf{D2Snap$_{.9,.6,.3}$} & 67.3 & \scriptsize [53.8, 78.5] & 84.05 & 21.01 & 3.51, 66.81 & 16.4 & 0.18 & 2.22 & 0.21 \\
        \hline
    \end{tabular*}
    \caption{Agent evaluation results across snapshot representations over $n = 52$ annotated snapshots, averaged across three runs at temperature $0$: agent task success rates with marginal 95\% Wilson intervals, which are per representation, not a basis for pairwise comparison. Mean byte and token sizes with token percentiles (5th, 95th; wide, as individual web page sizes vary). Mean model context utilisation (CU$_{\%}$, GPT-4o, $128 \times 10^3$ tokens). Mean snapshot size ratio against the mean raw DOM snapshot from the dataset ($\Delta_{\mathrm{Raw}}$). Mean latency normalised to the lowest recorded value (\cancel{GUI}$_{\mathrm{grounded}}$) with 1 SEM. Grounded GUI snapshots (GUI$_{\mathrm{grounded}}$) represent the baseline. 42\% of raw DOM snapshots exceeded the context window. Success with D2Snap-based representations is not significantly different from the baseline (e.g., D2Snap$_{.9,.3,.6}$, $+5.8$\%pt, McNemar, $p = 0.47$), nor with grounding text alone (\cancel{GUI}$_{\mathrm{grounded}}$, McNemar, $p = 0.37$).}
    \label{tab:3-1}
\end{table*}
\egroup

\begin{figure}[!t]
    \includegraphics[width=\linewidth]{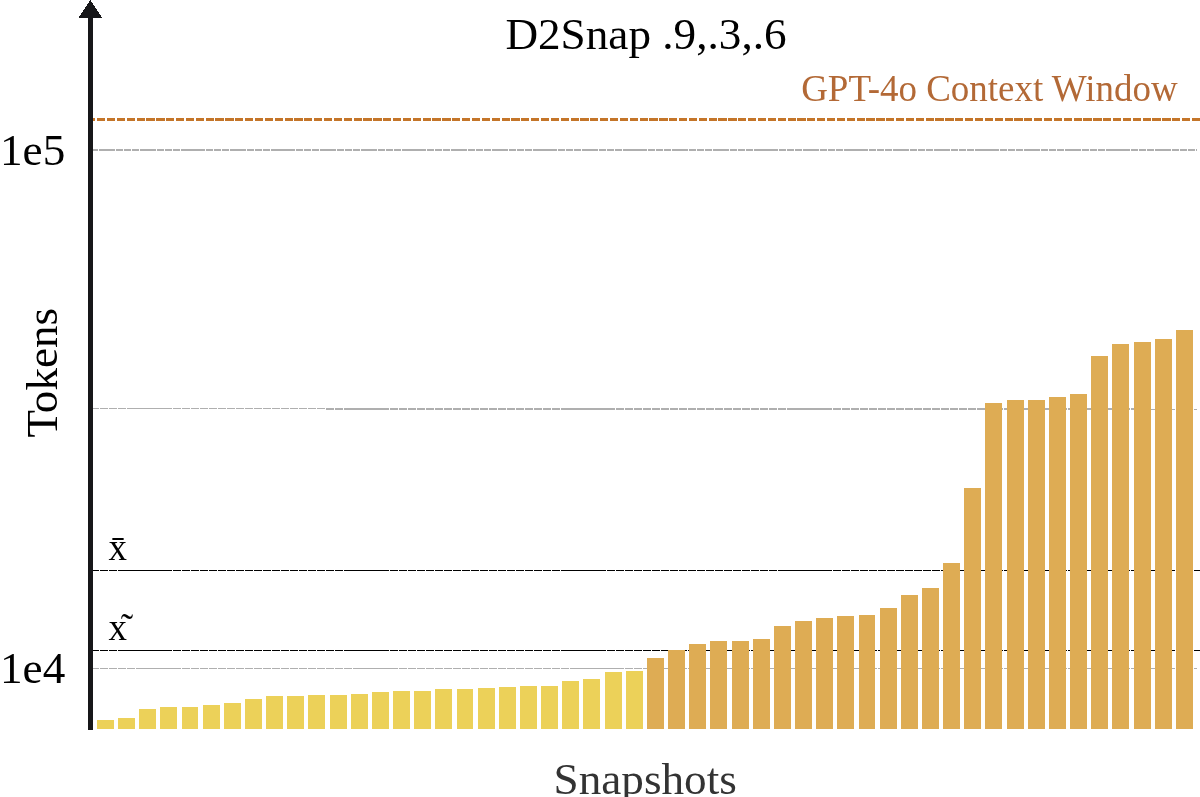}
    \caption{Ordered token sizes for each D2Snap-downsampled DOM from the dataset, produced by \textbf{D2Snap$_{.9,.3,.6}$}. All snapshots fit within the model's context window (GPT-4o, $1.28 \times 10^5$ tokens), at a mean utilisation of 16.5\% and a maximum of 54\%. 50\% are below $10^4$ tokens.}
    \label{fig:3-3}
\end{figure}

Raw DOM snapshots (\textbf{DOM}) exceed the model's context window in 42\% of the cases, at a mean CU$_{\%}$ of 109\%. Even at moderate downsampling ratios, every D2Snap-downsampled snapshot fits, and 13 of 15 assessed configurations have at most 20\% CU$_{\%}$ (\textbf{RQ~1})\footnote{We estimate LLM input tokens from serialised byte size. We map targets between original DOM and downsampled counterparts by assigning unique data attributes to the DOM. These are excluded from size measurements, as they stem from an opinionated mapping approach -- a source map would not affect snapshots directly. To not introduce bias, these attributes are applied to every element and values are incremental integers.}. At a CU$_{\%}$ of 16.5\%, our reference configuration \textbf{D2Snap$_{.9,.3,.6}$} achieves a success rate of 73\%, which is not significantly different from the baseline (\textbf{GUI$_{\mathrm{grounded}}$}) rate of 67\% ($+5.8$\%pt, 95\% CI $-13.6$ to $+26.0$\%pt; McNemar, $p = 0.47$) -- excluding a deficit (one-sided 95\%) beyond 11\%pt (\textbf{RQ~2}). Text-linearised representations (\textbf{D2Snap$_{1.0}$}, \textbf{D2Snap$_{\mathrm{MD}}$}) attain the lowest success rates among D2Snap configurations, underperforming our reference configuration significantly (42\%, $-30.8$\%pt, $p < 0.001$; 46\%, $-26.9$\%pt, $p = 0.002$). \tabref{tab:3-1} compares all evaluated snapshot representations on agent success, snapshot size, and LLM backend latency. \figref{fig:3-2} visualises success rate and token count among key representations. \figref{fig:3-3} plots token counts of every output of our reference D2Snap configuration.

At a success rate of 2\%, LLM vision capabilities evidently lack pixel-precision. Grounding is therefore necessary to make GUI snapshots effective (\textbf{GUI$_{\mathrm{grounded}}$}, 67\%). However, we detect no significant difference when using the grounding text without screenshots as standalone snapshots (\textbf{\cancel{GUI}$_{\mathrm{grounded}}$}); the agent attains a success rate of 62\% ($-5.8$\%pt, 95\% CI $-18.0$ to $+7.1$\%pt; McNemar, $p = 0.37$), which suggests that the text, rather than the image input, carries most utility-inherent features.

\paragraph{\mbox{Ablation Study.}} We infer the impact on snapshot utility of each D2Snap parameter by raising a single parameter to 0.9 (high) from an otherwise uniform 0.3 (low). Against the uniform \textbf{D2Snap$_{.3}$} reference (71\%, $\Delta_{\mathrm{Raw}} = 0.24$), this directionally yields: (A) $r_e$ -- low retention of elements (hierarchy) -- reduces size by 18\% at 73\%, so structural segregation can be traded almost freely; (B) $r_a$ -- low retention of attributes -- reduces size by 53\%, the largest saving of any parameter, but drops to 67\%, indicating that attributes add most to utility; and (C) $r_t$ -- low retention of text -- neither reduces size (1.9\%) nor changes success (71\%) noticeably. The embedded text consolidation technique (we deployed TextRank) narrows down to a semantic-to-verbosity optimisation problem. Notably, retaining HTML at low fidelity (\textbf{D2Snap$_{.6}$}, 73\%) outperforms full removal of HTML (\textbf{D2Snap$_{1.0}$}, 42\%; $+30.8$\%pt, McNemar, $p < 0.001$) -- supporting the use of DOM snapshots over abstract text-based snapshot representations. Retaining text at full length (\textbf{D2Snap$_{\mathrm{MD}}$}, 46\%) does not measurably change utility over discarding it (\textbf{D2Snap$_{1.0}$}; $+3.8$\%pt, $p = 0.32$), suggesting that natural language text verbosity contributes little to a DOM snapshot's utility.

\bgroup
\def\arraystretch{1.2}
\begin{table}[!t]\footnotesize
    \centering
    \begin{tabular*}{\columnwidth}{@{\extracolsep{\fill}}l r r r@{}}
        \hline
        & \multicolumn{3}{c}{\tiny$\overline{X}$ \footnotesize \textbf{Size}} \\
        & Tokens \scriptsize ($\times 10^3$) & $\Delta_{\mathrm{Raw}}$ & $\Delta$ \\
        \hline
        raw & 111.74 & 1 & -- \\
        \textbf{D2Snap$_{.1}$} & 30.73 & 0.275 & $-0.725$ \\
        \textbf{D2Snap$_{.2}$} & 28.26 & 0.253 & $-0.022$ \\
        \textbf{D2Snap$_{.3}$} & 26.07 & 0.238 & $-0.015$ \\
        \textbf{D2Snap$_{.4}$} & 25.29 & 0.226 & $-0.012$ \\
        \textbf{D2Snap$_{.5}$} & 23.58 & 0.208 & $-0.018$ \\
        \textbf{D2Snap$_{.6}$} & 23.32 & 0.202 & $-0.006$ \\
        \textbf{D2Snap$_{.7}$} & 21.37 & 0.183 & $-0.019$ \\
        \textbf{D2Snap$_{.8}$} & 11.50 & 0.115 & $-0.068$ \\
        \textbf{D2Snap$_{.9}$} & 10.62 & 0.109 & $-0.006$ \\
        \textbf{D2Snap$_{1.0}$} & 4.69 & 0.057 & $-0.052$ \\
        \hline
    \end{tabular*}
    \caption{Mean token counts and size ratio against the mean raw DOM snapshot from the dataset ($\Delta_{\mathrm{Raw}}$) over uniform D2Snap parameter configurations $r = r_e = r_a = r_t \in \{0.1, \ldots, 1.0\}$, supplemented with absolute change ($\Delta$). The progression is strictly monotonic (Spearman $\rho = -1.000$). Over the operating range $r$, it fits $\hat{\Delta}_{\mathrm{Raw}} = 0.311 - 0.226\,r$ (Pearson correlation $-0.965$, $R^2 = 0.93$). The drop from $0.7$ to $0.8$ is due to the elimination of the high-frequency \texttt{class} attribute.}
    \label{tab:3-2}
\end{table}
\egroup

\begin{figure}[!t]
    \includegraphics[width=\linewidth]{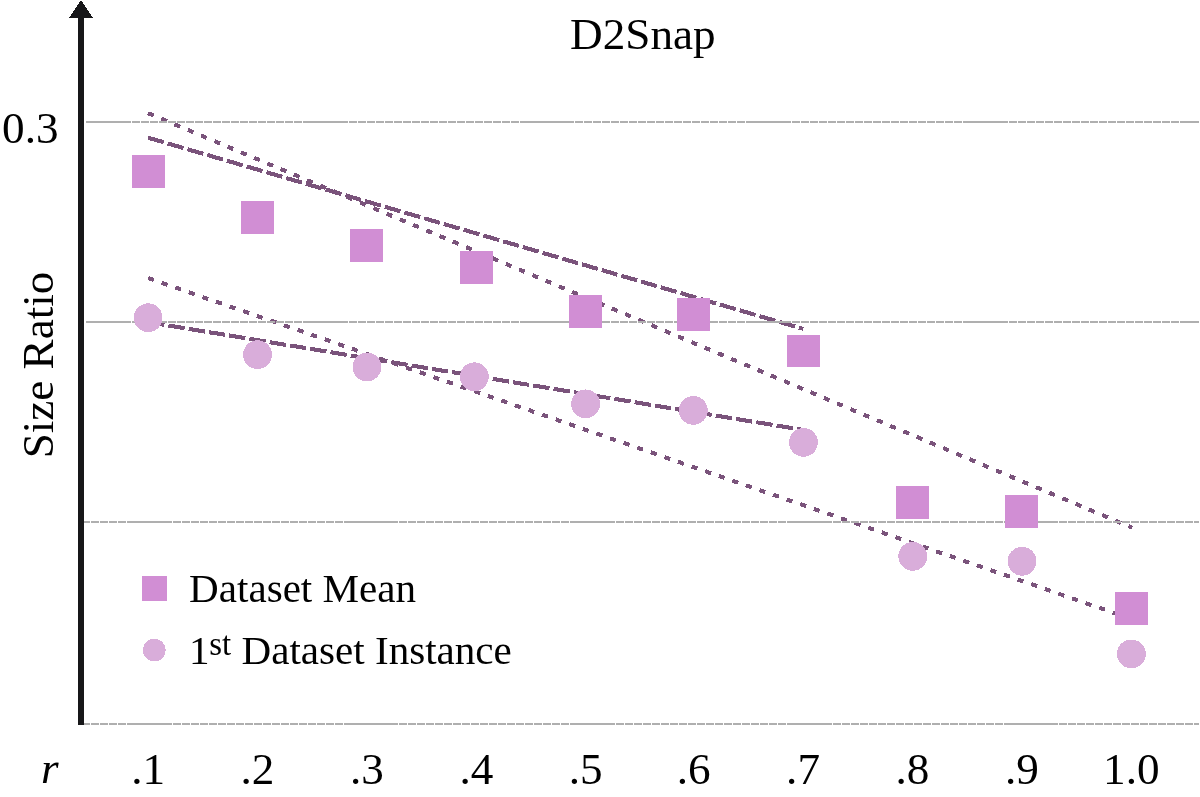}
    \caption{Mean size ratio against the mean raw DOM snapshot from the dataset ($\Delta_{\mathrm{Raw}}$) over uniform D2Snap parameter configurations $r = r_e = r_a = r_t \in \{0.1, \ldots, 1.0\}$. The progression is strictly monotonic (Spearman $\rho = -1.000$). Over the operating range $r$, it fits $\hat{\Delta}_{\mathrm{Raw}} = 0.311 - 0.226\,r$ (Pearson correlation $-0.965$, $R^2 = 0.93$). The drop from $0.7$ to $0.8$ is due to the elimination of the high-frequency \texttt{class} attribute. The size progression on the first snapshot from the dataset resembles the trend.}
    \label{fig:3-4}
\end{figure}

\paragraph{\mbox{Range Study.}} D2Snap's output ranges from the original HTML (\textbf{D2Snap$_0$}) to fully flattened content -- virtually Markdown with inlined actionable elements (\textbf{D2Snap$_{1.0}$}). We record the uniform configurations $r = r_e = r_a = r_t \in \{0.1, \ldots, 1.0\}$. Size reduction is strictly monotonic across this range (Spearman $\rho = -1.000$), satisfying our strong aim. It further approximates a constant reduction rate: a linear regression of the mean size ratio against $r$ over the operating range fits $\hat{\Delta}_{\mathrm{Raw}} = 0.311 - 0.226\,r$ (Pearson correlation $-0.965$, $R^2 = 0.93$), satisfying our weak aim. The drop from $0.7$ to $0.8$ stands out: it is due to the \texttt{class} attribute~\cite{WHATWG_1}, which our experimental attribute scoring places at $0.77$ and which is therefore discarded once $r_a$ exceeds that score. Aligned with prior expectation, this is a high-frequency attribute. Restricting the regression to $r \in \{0.1, \ldots, 0.7\}$, below that threshold, yields a closer fit of $\hat{\Delta}_{\mathrm{Raw}} = 0.285 - 0.146\,r$ (Pearson correlation $-0.993$, $R^2 = 0.99$). \tabref{tab:3-2} reports mean snapshot token counts and size ratios across the recorded range; \figref{fig:3-4} plots the size ratios against the regressions.
\section{Related Work}

\paragraph{\mbox{Context Engineering.}} Input token efficiency has been approached with increasing web agent specificity: LLMLingua~\cite{JiangWu2023, JiangWu2024, PanWu2024} discards low-saliency tokens by perplexity in continuous natural language prose -- a noteworthy alternative to our TextRank component of D2Snap. Set-of-Mark prompting augments a GUI snapshot with visual cues~\cite{YangZhang2023}, which has influenced state-of-the-art production agents such as \textit{Browser Use}~\cite{BrowserUse_1}. Token-extraction and -stripping approaches reduce HTML directly, yet apply uniform transformations that disregard differential DOM node semantics~\cite{ChangKayed2006, GuptaKaiser2003}. Closest to our work, structured-text representations serialise the page directly: most obviously, the accessibility tree~\cite{BoxhallGash2016} derived from the DOM~\cite{ZhouXu2024}. Markdown constitutes an orthogonal representation~\cite{Ango2025}. Such representations typically follow element extraction -- surfacing a designated subset of elements for translation. Element extraction has built on above-outlined transformations, but also on LLM inference, such as relevance ranking or predictors conditioned on prior actions~\cite{GurFuruta2024, DengGu2023} -- at the cost of added latency. In contrast, D2Snap downsamples the full DOM under an explicit fidelity-size trade-off, preserving features that constitute or discriminate action targets across the entire representation at a controllable ratio, and emitting a valid DOM.

\paragraph{\mbox{Web Agents.}} A growing body of work has proposed architectures for generalist, LLM-based web agents~\cite{KimBaldi2023, GurFuruta2023, ZhouXu2024, HeYao2024, ZhengGou2024, BrowserUse_0, OpenAI_7} -- based on DOM, GUI or hybrid snapshots. A parallel line of work has focused on web agent evaluation: Mind2Web~\cite{DengGu2023} introduced a static dataset of web browsing tasks to assess web agent success. Online-Mind2Web~\cite{XueQi2025} is a smaller-scale derivative targeting execution on the live web; WebVoyager~\cite{HeYao2024} likewise targets the live web. These datasets are designed to capture actual web browsing journeys rather than final results~\cite{OSUNLP_1}. The evaluation of snapshot utility in isolation, however, has received comparatively little attention -- generalist web agent evaluation results merely allow indirect assessment of snapshot representations. Our work closes this gap, presenting a static dataset to assess different snapshot representations. Our related experimental framework is designed to assess the impact of a specific snapshot representation independently of individual web agent logic. It does, moreover, not require human or AI judgement (e.g., WebJudge~\cite{OSUNLP_2}\footnote{We utilised WebJudge once to validate task solution UI state trajectories in our dataset. It is not used in our repeatable experimental framework.}), as it defines objective rules to determine success.
\section{Conclusion}

DOM snapshots -- if not for their excessive LLM input footprint -- represent a compelling alternative to GUI snapshots. In this paper, we introduced DOM downsampling: trading DOM fidelity for size, whilst preserving a valid DOM.

\subsection{Contributions}

We propose D2Snap -- an algorithm to downsample the DOM, premised on preserving actionability and actionability-discriminating features, intended as a size-reducing pre-processing technique for DOM snapshots used with web agents. On D2Snap-downsampled DOM snapshots, a minimal, snapshot-variant web agent (GPT-4o, context size: $128\times 10^3$ tokens) achieves a web browsing task success rate that is not significantly different from that of a grounded GUI snapshot baseline ($+5.8$\%pt, 95\% CI $-13.6$ to $+26.0$\%pt), excluding a deficit (one-sided 95\%) beyond 11\%pt, at a mean model context utilisation of 16.5\%. Every D2Snap-downsampled snapshot fits within the context window; 42\% of raw DOM snapshots exceed it. We further detect no significant difference in success between grounded GUI snapshots and the grounding text alone ($-5.8$\%pt, 95\% CI $-18.0$ to $+7.1$\%pt), which suggests that image input contributes little to snapshot utility. Ablation on success indicates that retention of attributes contributes most to snapshot utility. In turn, low retention of element hierarchy is sufficient (73\%), yet full removal of HTML is not (46\%). We show that D2Snap downsamples strictly monotonically and approximately linearly across its operating range (linear regression: slope $-0.226$, $R^2 = 0.93$).

The data and code repository associated with this paper is publicly available at \url{github.com/webfuse-com/D2Snap}, mirrored at \url{github.com/t-ski/D2Snap}.

\subsection{Limitations}

\paragraph{Evaluation.} Our evaluation dataset has a modest size ($n = 52$), which limits the precision of our estimates. Our reference configuration was selected among the results. It ties with three further configurations at the highest observed success rate, and among those exhibits the lowest context utilisation -- a criterion independent of success. Our evaluation relies exclusively on GPT-4o, so results should be read as indicative of snapshot utility for few real-world-representative web pages and a highly capable model. Elicited once, our attribute scoring is best regarded as a model-informed prior.

\paragraph{Integration.} Web browsers strictly constrain DOM-access to cross-origin documents embedded via the \texttt{iframe} element~\cite{WHATWG_0, WHATWG_1}. Whilst GUI snapshots are able to capture what is visually rendered within embedded documents' viewports~\cite{Playwright_1}, DOM snapshots are not. The same limitation holds for graphics-based applications via the \texttt{canvas} element~\cite{WHATWG_1}.

\subsection{Future Work}

\paragraph{Evaluation.} Automated, systematic exploration of the downsampling parameter space -- across a broader range of frontier LLMs and web applications -- would identify optimal configurations and establish generalisability of the approach. Moreover, we elicited attribute scoring heuristics from a single model under a generic eliciting prompt; future work could investigate alternative elicitation strategies.

\paragraph{Integration.} Since D2Snap preserves a valid DOM, it can serve as an intermediate size control measure within a snapshot pipeline. In a subsequent processing step, for instance, the downsampled DOM could be translated to an accessibility tree. In a preceding step, certain subtrees could be pruned, e.g., due to invisibility or redundancy.

\paragraph{\mbox{Beyond Web Agents.}} The applicability of DOM downsampling is not limited to the web agent domain. Other domains that might benefit from the concept include accessibility, content summarisation, and UI testing -- all of which rely on the DOM.

    \newpage
    \small
    \hypersetup{urlcolor=black}
    \setlength{\bibsep}{0.5ex}
    \setlength{\bibhang}{1.5em}
    \bibliographystyle{plainnat}
    \bibliography{_references.bib}

    \appendix
    \onecolumn
    \raggedbottom
    \hypersetup{urlcolor=mediumred}
    \lstset{basicstyle=\ttfamily\scriptsize\linespread{1.1}\selectfont}
    \normalsize
    \clearpage
\section{GUI Snapshot Examples}
\label{att:a}

\subsection{GUI Snapshot}

\noindent{\small A GUI snapshot corresponds to a screenshot.}

\smallskip

\begin{llminput}[title={}, width=.5225\linewidth]
\includegraphics[width=\linewidth]{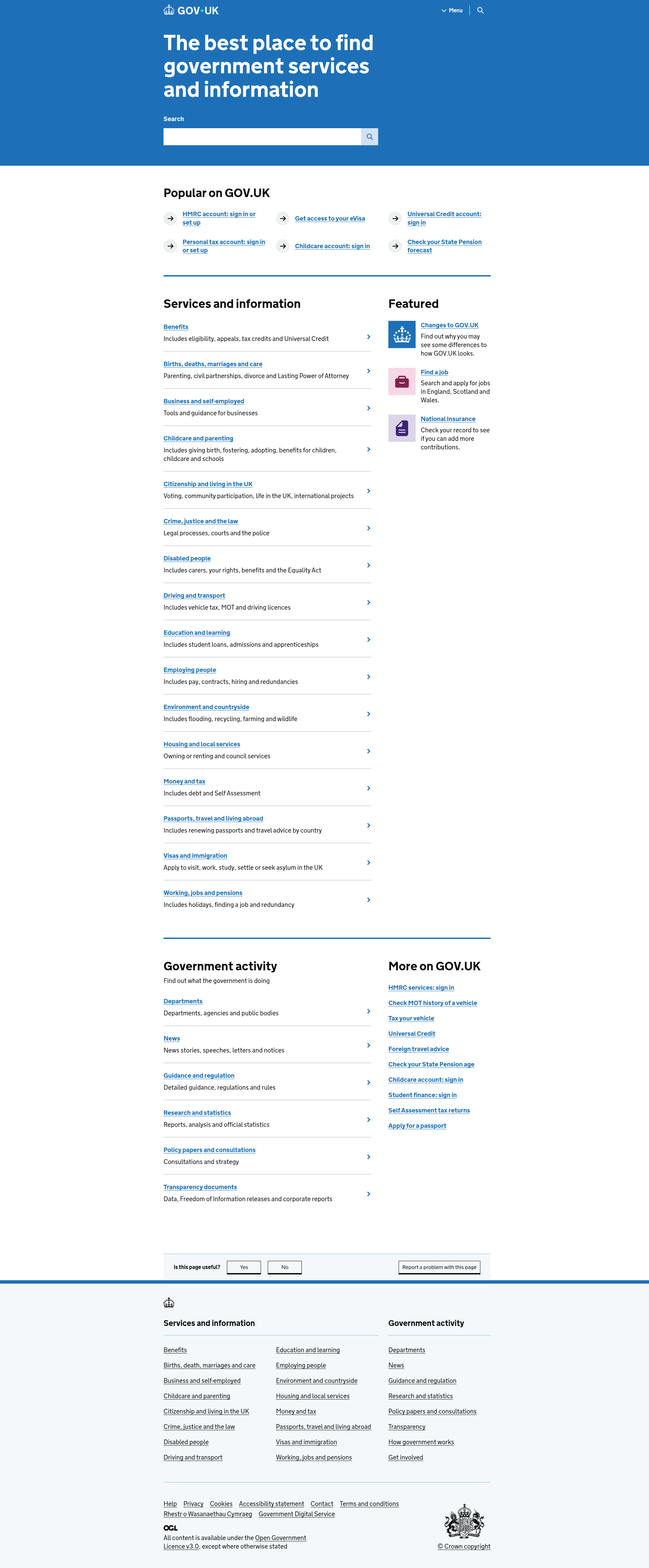}
\end{llminput}

\newpage

\subsection{Grounded GUI Snapshot}

\noindent{\small A grounded GUI snapshot corresponds to a GUI snapshot (image) enhanced with visual cues -- to bridge the lack of LLM vision pixel-precision. Visual cues are coloured bounding boxes identified by numerical indices (Set-of-Mark). The grounding (text) describes actionable elements associated with these identifiers providing (tag) name and text contents.}

\smallskip

\begin{llminput}[title={}]
\begin{minipage}[!t]{.5\linewidth}
\includegraphics[width=\linewidth]{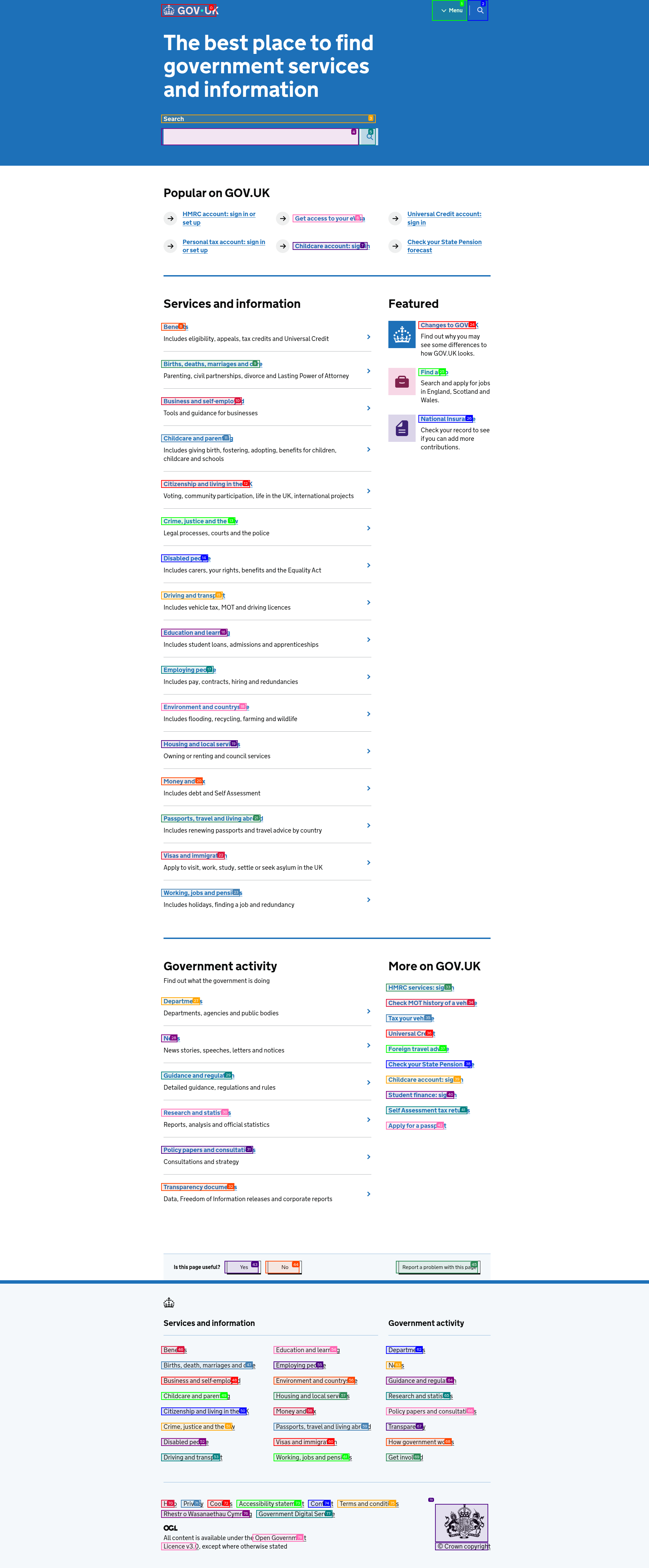}
\end{minipage}
\hfill
\begin{minipage}[!t]{.45\linewidth}
\fontsize{6}{7}\selectfont
\begin{alltt}
[0] A ""
[1] BUTTON "Menu"
[2] BUTTON "Search GOV.UK"
[3] LABEL "Search"
[4] INPUT ""
[5] BUTTON "Search GOV.UK"
[6] A "Get access to your eVisa"
[7] A "Childcare account: sign in"
[8] A "Benefits"
[9] A "Births, deaths, marriages and care"
[10] A "Business and self-employed"
[11] A "Childcare and parenting"
[12] A "Citizenship and living in the UK"
[13] A "Crime, justice and the law"
[14] A "Disabled people"
[15] A "Driving and transport"
[16] A "Education and learning"
[17] A "Employing people"
[18] A "Environment and countryside"
[19] A "Housing and local services"
[20] A "Money and tax"
[21] A "Passports, travel and living abroad"
[22] A "Visas and immigration"
[23] A "Working, jobs and pensions"
[24] A "Changes to GOV.UK"
[25] A "Find a job"
[26] A "National Insurance"
[27] A "Departments"
[28] A "News"
[29] A "Guidance and regulation"
[30] A "Research and statistics"
[31] A "Policy papers and consultations"
[32] A "Transparency documents"
[33] A "HMRC services: sign in"
[34] A "Check MOT history of a vehicle"
[35] A "Tax your vehicle"
[36] A "Universal Credit"
[37] A "Foreign travel advice"
[38] A "Check your State Pension age"
[39] A "Childcare account: sign in"
[40] A "Student finance: sign in"
[41] A "Self Assessment tax returns"
[42] A "Apply for a passport"
[43] BUTTON "Yes this page is useful"
[44] BUTTON "No this page is not useful"
[45] BUTTON "Report a problem with this page"
[46] A "Benefits"
[47] A "Births, death, marriages and care"
[48] A "Business and self-employed"
[49] A "Childcare and parenting"
[50] A "Citizenship and living in the UK"
[51] A "Crime, justice and the law"
[52] A "Disabled people"
[53] A "Driving and transport"
[54] A "Education and learning"
[55] A "Employing people"
[56] A "Environment and countryside"
[57] A "Housing and local services"
[58] A "Money and tax"
[59] A "Passports, travel and living abroad"
[60] A "Visas and immigration"
[61] A "Working, jobs and pensions"
[62] A "Departments"
[63] A "News"
[64] A "Guidance and regulation"
[65] A "Research and statistics"
[66] A "Policy papers and consultations"
[67] A "Transparency"
[68] A "How government works"
[69] A "Get involved"
[70] A "Help"
[71] A "Privacy"
[72] A "Cookies"
[73] A "Accessibility statement"
[74] A "Contact"
[75] A "Terms and conditions"
[76] A "Rhestr o Wasanaethau Cymraeg"
[77] A "Government Digital Service"
[78] A "Open Government Licence v3.0"
[79] A "$\copyright$ Crown copyright"
\end{alltt}
\end{minipage}
\end{llminput}
\clearpage
\section{Example DOM Snapshots}
\label{att:b}

\begin{attbox}
{\scriptsize\texttt{pizza.html}}

\begin{lstlisting}[aboveskip=0.5em,belowskip=0em]
<section class="container" tabindex="3" required="true" type="example">
  <div class="mx-auto" data-topic="products" required="false">
    <h1>Our Pizza</h1>
    <div>
      <div class="shadow-lg">
        <h2>Margherita</h2>
        <p>
         A simple classic: mozzarella, tomatoes and basil.
         An everyday choice!
        </p>
        <button type="button">Add</button>
      </div>
      <div class="shadow-lg">
        <h2>Capricciosa</h2>
        <p>
          A rich taste: mozzarella, ham, mushrooms, artichokes and olives.
          A true favourite!
        </p>
        <button type="button">Add</button>
      </div>
    </div>
  </div>
</section>
\end{lstlisting}
\end{attbox}

\begin{attbox}
{\scriptsize$D2Snap(\texttt{pizza.html})\ |\ r_e = r_a = r_t = 0.3$ \footnotesize{($\Delta_{\mathrm{Raw}}~68\%$)}}

\begin{lstlisting}[aboveskip=0.5em,belowskip=0em]
<section class="container" required="true" type="example">
  <div class="mx-auto" required="false">
    # Our Pizza
    <div>
      ## Margherita
      A simple classic: mozzarella, tomatoes and basil.
      An everyday choice!
      <button type="button">Add</button>
      ## Capricciosa
      A rich taste: mozzarella, ham, mushrooms, artichokes and olives.
      A true favourite!
      <button type="button">Add</button>
    </div>
  </div>
</section>
\end{lstlisting}
\end{attbox}

\begin{attbox}
{\scriptsize$D2Snap(\texttt{pizza.html})\ |\ r_e = 0.9, r_a = 0.3, r_t = 0.6$ \footnotesize{($\Delta_{\mathrm{Raw}}~46\%$)}}

\begin{lstlisting}[aboveskip=0.5em,belowskip=0em]
<section class="container" required="true" type="example">
  # Our Pizza
  ## Margherita
  A simple classic: mozzarella, tomatoes and basil.
  <button type="button">Add</button>
  ## Capricciosa
  A rich taste: mozzarella, ham, mushrooms, artichokes and olives.
  <button type="button">Add</button>
</section>
\end{lstlisting}
\end{attbox}

\begin{attbox}
{\scriptsize$D2Snap(\texttt{pizza.html})\ |\ r_e = r_a = 1, r_t = 0$ \footnotesize{($\Delta_{\mathrm{Raw}}~35\%$)}}

\begin{lstlisting}[aboveskip=0.5em,belowskip=0em]
# Our Pizza
## Margherita
A simple classic: mozzarella, tomatoes and basil.
An everyday choice!
<button>Add</button>
## Capricciosa
A rich taste: mozzarella, ham, mushrooms, artichokes and olives.
A true favourite!
<button>Add</button>
\end{lstlisting}
\end{attbox}
\clearpage
\section{LLM-Elicited$^1$ Attribute Scoring}
\label{att:c}

\begin{attbox}
\begin{llminput}
For each standard HTML attribute: Score its UI-features, i.e., its semantics in the inherent user interface (`S'), using a ratio scale on the unit interval $[0, 1]$. Subsequently calibrate these scores against the global frequency of each attribute so that, in a DOM representative of an average document, the fraction of attribute occurrences with a score of at most `s' is approximately s, for every s in $[0, 1]$. Respond with a list of tuples (attribute, S).
\end{llminput}

{
\setlength{\columnsep}{1em}
\begin{multicols}{2}
\begin{llmoutput}
\begin{alltt}
('accept', 0.24)
('accept-charset', 0.01)
('accesskey', 0.00)
('action', 0.27)
('alt', 0.46)
('aria-*', 0.62)
('autocapitalize', 0.01)
('autocomplete', 0.15)
('autofocus', 0.16)
('autoplay', 0.03)
('checked', 0.36)
('cite', 0.01)
('class', 0.77)
('content', 0.05)
('contenteditable', 0.15)
('controls', 0.36)
('coords', 0.00)
('datetime', 0.15)
('disabled', 0.34)
('download', 0.02)
('draggable', 0.01)
('dropzone', 0.00)
('enctype', 0.17)
('for', 0.30)
('form', 0.26)
('hidden', 0.28)
('href', 0.91)
('hreflang', 0.02)
('http-equiv', 0.01)
('id', 0.68)
('ismap', 0.00)
('label', 0.31)
('lang', 0.30)
('loop', 0.04)
('max', 0.23)
\end{alltt}
\end{llmoutput}

\begin{llmoutput}[title={}]
\begin{alltt}
('maxlength', 0.21)
('media', 0.03)
('method', 0.27)
('min', 0.23)
('minlength', 0.21)
('multiple', 0.26)
('muted', 0.04)
('name', 0.40)
('novalidate', 0.02)
('pattern', 0.22)
('ping', 0.00)
('placeholder', 0.25)
('poster', 0.03)
('preload', 0.02)
('readonly', 0.33)
('rel', 0.18)
('required', 0.33)
('role', 0.32)
('selected', 0.35)
('shape', 0.00)
('size', 0.04)
('sizes', 0.19)
('spellcheck', 0.02)
('src', 0.83)
('srcset', 0.20)
('step', 0.22)
('style', 0.10)
('tabindex', 0.16)
('target', 0.17)
('title', 0.29)
('type', 0.38)
('usemap', 0.01)
('value', 0.43)
('wrap', 0.01)

('*', 0.00) [ed.]²
\end{alltt}
\end{llmoutput}
\end{multicols}
}
\end{attbox}

{\fontsize{8}{10}\selectfont
\medskip
$^1$ GPT-4o (\texttt{gpt-4o-2024-11-20}~\cite{OpenAI_4})\par
$^2$ Manual edit by the author.\par
}
\clearpage
\section{D2Snap}
\label{att:d}

\par\noindent
\begin{minipage}{0.7\textwidth}
\begin{lstlisting}[aboveskip=0em,belowskip=0em,language=pseudocode,mathescape,basicstyle=\fontsize{8}{16}]
INPUT: ELEMENT_NODE $\equiv$ DOM, $r_e$, $r_a$, $r_t$ $\in$ $[0, 1]$, $\mathcal{A}$
OUTPUT: ELEMENT_NODE $\equiv$ DOM
PROCEDURE D2Snap $\in$ $\mathcal{O}$(|\text{DOM}|)$^{\S}$:
  // POST-ORDER DOM TRAVERSAL:
  for CHILD_ELEMENT_NODE of ELEMENT_NODE$_{\mathrm{Children}}$:
    D2Snap(CHILD_ELEMENT_NODE, $r_e$, $r_a$, $r_t$, $\mathcal{A}$)

  // TEXT NODE SUB-PROCEDURE:
  for TEXT_NODE of ELEMENT_NODE$_{\mathrm{Text}}$:
    SENTENCES $\leftarrow$ Tokenize(TEXT_NODE$_{\mathrm{TextContent}}$)
    SENTENCES$_{\mathrm{ranked}}$ $\leftarrow$ TextRank$_{\textrm{Sentence}}$(SENTENCES)
    SENTENCES$_{\mathrm{selected}}$ $\leftarrow$ SENTENCES$_{\mathrm{ranked}}$[0 : $\max(\lceil(1 - r_t) \cdot |\text{SENTENCES}|\rceil, 1)$]
    SENTENCES$_{\mathrm{re-ordered}}$ $\leftarrow$ Order$_{\textrm{Natural}}$(SENTENCES$_{\mathrm{selected}}$)
    TEXT_NODE$_{\mathrm{TextContent}}$ $\leftarrow$ Join(SENTENCES$_{\mathrm{re-ordered}}$)

  // ATTRIBUTE NODE SUB-PROCEDURE:
  for ATTRIBUTE_NODE of ELEMENT_NODE$_{\mathrm{Attributes}}$:
    if $\mathcal{A}$(ATTRIBUTE_NODE$_{\mathrm{Name}}$) < $r_a$:
      Remove(ATTRIBUTE_NODE)

  // ELEMENT NODE SUB-PROCEDURE:
  switch Class(ELEMENT_NODE):
    case Class.ACTIONABLE:
      break
    case Class.TEXT_FORMATTING:
      TEXT_NODE $\leftarrow$ Markdown(ELEMENT_NODE)
      Replace(ELEMENT_NODE, TEXT_NODE)
      break
    default:
      d $\leftarrow$ Depth$_{\mathrm{frozen}}$(ELEMENT_NODE)
      if d > 1 and $\lfloor \text{d} \cdot r_e \rfloor$ > $\lfloor (\text{d} - 1) \cdot r_e \rfloor$:
        Replace(ELEMENT_NODE, ELEMENT$_{\mathrm{Children}}$)
        break

  // RETURN REFERENCE TO MUTATED DOM:
  return ELEMENT_NODE
\end{lstlisting}

{\fontsize{8}{10}\selectfont
\bigskip
$^\S$ We presume that \textit{TextRank}, which is embedded in D2Snap's text sub-procedure, is bounded by sentence count and number of iterations to convergence.\par
}
\end{minipage}
\clearpage
\section{Annotated Dataset Entry}
\label{att:e}

\begin{lstlisting}[label={lst:d-1}]
{
  "id": "govuk-0",
  "trajectories": [
    [
      {
        "css_selector": "li.gem-c-cards__list-item:nth-of-type(15) > div.gem-c-cards__list-item-wrapper > h3.gem-c-cards__sub-heading
          .govuk-heading-s > a.govuk-link.gem-c-cards__link.gem-c-force-print-link-styles",
        "bounding_box": [ [ 480, 2473 ], [ 609, 108 ] ],
        "bu_identifier": 22
      }
    ],
    [
      {
        "css_selector": "li.govuk-footer__list-item:nth-of-type(15) > a.govuk-footer__link",
        "bounding_box": [ [ 810, 4222 ], [ 179, 18 ] ],
        "bu_identifier": 60
      }
    ],
    [
      {
        "css_selector": "#search-main-ec8e678d",
        "bounding_box": [
          [ 480, 376 ], [ 581, 50 ] ],
        "bu_identifier": 4
      },
      {
        "css_selector": "div.gem-c-search.govuk-\\!-display-none-print.gem-c-search--large.gem-c-search--homepage
          .gem-c-search--on-govuk-blue.gem-c-search--separate-label.govuk-\\!-margin-bottom-0 > div.gem-c-search__item-wrapper > div
          .gem-c-search__item.gem-c-search__submit-wrapper:nth-of-type(2) > button.gem-c-search__submit",
        "bounding_box": [
          [ 1060, 376 ],
          [ 50, 50 ]
        ],
        "bu_identifier": 5
      }
    ],
    [
      {
        "css_selector": "li.gem-c-layout-super-navigation-header__dropdown-list-item:nth-of-type(15) > a.govuk-link
          .gem-c-layout-super-navigation-header__navigation-second-item-link"
      }
    ],
    [
      {
        "css_selector": "a.gem-c-layout-super-navigation-header__search-item-link"
      },
      {
        "css_selector": "div.gem-c-search.govuk-\\!-display-none-print.gem-c-search--large.gem-c-search--on-white
          .gem-c-search--separate-label.govuk-\\!-margin-bottom-0 > div.gem-c-search__item-wrapper > div
          .gem-c-search__item.gem-c-search__submit-wrapper:nth-of-type(2) > button.gem-c-search__submit"
      }
    ]
  ]
}
\end{lstlisting}
\clearpage
\section{Evaluation System Prompt}
\label{att:f}
\hyphenpenalty=0
\exhyphenpenalty=0
\lstdefinelanguage{systemprompt}{
    morekeywords={
        SNAPSHOT_DESCRIPTION,
        SCHEMA_DESCRIPTION,
        EXAMPLE_SNAPSHOT, EXAMPLE_RESPONSE
    }
}

\begin{adjustbox}{minipage=0.7\textwidth}
    {\small To control the system prompt used in our evaluation, we define a generic template. The template contains variable tags (\codeinline{\{\{ <TAG> \}\}}) for each snapshot specific part of the prompt. Below are the template, and specific substitutes per snapshot class.}
\end{adjustbox}

\subsection{Template}

\begin{llminput}[title={}]
\begin{alltt}
# Identity

You are an AI agent that solves web-based tasks on behalf of a human user.

> Assume that today is July 16, 2025.

# Instructions

The user provides you with a web-based task, and serialised state of the web application to solve the task with.
A task may be iterative, so it may not be possible to solve the task completely, but only partially with the given state.

Based on the state representation, your goal is to suggest all elements required to interact with in order to solve the task.
It is important that the list of elements corresponds to a complete interaction trajectory. High precision when
referencing target elements is key in order to be able to reproduce the interactions on the respective user interface.

## Input

### Task

The web-based task is denoted with the prefix `TASK:`, e.g. "TASK: Show 4-star hotels in Amsterdam".

### Snapshot

\textbf{\{\{ SNAPSHOT_DESCRIPTION \}\}}

# Output

Follow these rules when considering an element for interaction:

- In case there are multiple trajectories to solve the task, rank them in memory according to human-readability
  and choose the highest ranked alternative
- If there are alternative elements per trajectory which seem to do the same thing, choose the most expressive alternative 
- Suppose there are only point and click actions, so never imply any other interaction
- Try not to over-suggest, i.e., suggest only mandatory elements to solve the task with

## Schema

\textbf{\{\{ SCHEMA_DESCRIPTION \}\}}

# Examples

Consider the web-based task "TASK: Calculate the sum of 2 and 3.".

<user_query>
  TASK: Calculate the sum of 2 and 3.
</user_query>

<user_query>
  \textbf{\{\{ EXAMPLE_SNAPSHOT \}\}}
</user_query>

<assistant_response>
  \textbf{\{\{ EXAMPLE_RESPONSE \}\}}
</assistant_response>
\end{alltt}
\end{llminput}

\newpage
\subsection{GUI}

\begin{attbox}
\footnotesize{\codeinline{\{\{ SNAPSHOT_DESCRIPTION \}\}}}

\begin{llminput}[title={}]
\begin{alltt}
You are provided with a screenshot, namely the rendered GUI.
\end{alltt}
\end{llminput}
\end{attbox}

\begin{attbox}
\footnotesize{\codeinline{\{\{ SCHEMA_DESCRIPTION \}\}}}

\begin{llminput}[title={}]
\begin{alltt}
Target elements by their spatial center pixel coordinates. This means, refer to
them through an x (horizontal) and a y (vertical) pixel coordinate relative to
the origin, which is in the top left corner of the image.
\end{alltt}
\end{llminput}
\end{attbox}

\begin{attbox}
\footnotesize{\codeinline{\{\{ EXAMPLE_SNAPSHOT \}\}}\footnote{
    \scriptsize\url{https://github.com/webfuse-com/D2Snap/blob/main/.github/gui.b64-example.png}
}}

\begin{llminput}[title={}]
\begin{alltt}
``` base64
data:image/png;base64,iVBORw0KGgoAAAANSUhEUgAAAMgAAACWCAMAAACs...
```
\end{alltt}
\end{llminput}
\end{attbox}

\begin{attbox}
\footnotesize{\codeinline{\{\{ EXAMPLE_RESPONSE \}\}}}

\begin{llminput}[title={}]
\begin{alltt}
``` json
[
  \{
    "elementDescription": "Field that contains the mathematical expression to
      be solved.",
    "x": 100,
    "y": 47
  \},
  \{
  "elementDescription": "Button that triggers the calculation of the provided
    mathematical expression.",
    "x": 100,
    "y": 197
  \}
]
```
\end{alltt}
\end{llminput}
\end{attbox}

\newpage
\subsection{Grounded GUI}

\begin{attbox}
\footnotesize{\codeinline{\{\{ SNAPSHOT_DESCRIPTION \}\}}}

\begin{llminput}[title={}]
\begin{alltt}
You are provided with two means of input:

1. A screenshot of the browser with bounding boxes and related numeric identifiers.
2. A list of interactive elements with format `[index] type "text"`. `index` is the
numeric identifier, `type` is an HTML element type (button, input, etc.), and
`text` is the element description.

> Numeric identifiers across means of input are consistent.
\end{alltt}
\end{llminput}
\end{attbox}

\begin{attbox}
\footnotesize{\codeinline{\{\{ SCHEMA_DESCRIPTION \}\}}}

\begin{llminput}[title={}]
\begin{alltt}
Target elements by their numeric identifiers as given across both means of input.
\end{alltt}
\end{llminput}
\end{attbox}

\begin{attbox}
\footnotesize{\codeinline{\{\{ EXAMPLE_SNAPSHOT \}\}}\footnote{
    \scriptsize\url{https://github.com/webfuse-com/D2Snap/blob/main/.github/gui.b64-example.bu.png}}
}

\begin{llminput}[title={}]
\begin{alltt}
``` base64
data:image/png;base64,iVBORw0KGgoAAAANSUhEUgAAAMgAAACWCAIAAAAU...
```
\end{alltt}
\end{llminput}

\begin{llminput}[title={}]
\begin{alltt}
```
[0] INPUT "Type expression"
[1] BUTTON "Solve"
[2] A ""
```
\end{alltt}
\end{llminput}
\end{attbox}

\begin{attbox}
\footnotesize{\codeinline{\{\{ EXAMPLE_RESPONSE \}\}}}

\begin{llminput}[title={}]
\begin{alltt}
``` json
[
  \{
    "elementDescription": "Field that contains the mathematical expression to
      be solved.",
    "identifier": 0
  \},
  \{
    "elementDescription": "Button that triggers the calculation of the provided
      mathematical expression.",
    "identifier": 1
  \}
]
```
\end{alltt}
\end{llminput}
\end{attbox}

\newpage
\subsection{DOM/D2Snap}

\begin{attbox}
\footnotesize{\codeinline{\{\{ SNAPSHOT_DESCRIPTION \}\}}}

\begin{llminput}[title={}]
\begin{alltt}
You are provided with HTML, namely a serialised DOM.
\end{alltt}
\end{llminput}
\end{attbox}

\begin{attbox}
\footnotesize{\codeinline{\{\{ SCHEMA_DESCRIPTION \}\}}}

\begin{llminput}[title={}]
\begin{alltt}
Target elements by shortest unique CSS selector, e.g. `#app > button:first-child`.
\end{alltt}
\end{llminput}
\end{attbox}

\begin{attbox}
\footnotesize{\codeinline{\{\{ EXAMPLE_SNAPSHOT \}\}}}

\begin{llminput}[title={}]
\begin{alltt}
``` html
<main>
  <h1>Calculator</h1>
  <input id="expression" class="field" type="text" placeholder="3 * 4">
  <button id="submit" type="button">Solve</button>
  <div id="result">
    <span></span>
  </div>
```
\end{alltt}
\end{llminput}
\end{attbox}

\begin{attbox}
\footnotesize{\codeinline{\{\{ EXAMPLE_RESPONSE \}\}}}

\begin{llminput}[title={}]
\begin{alltt}
``` json
[
  \{
    "elementDescription": "Field that contains the mathematical expression to
      be solved.",
    "cssSelector": "#expression"
  \},
  \{
    "elementDescription": "Button that triggers the calculation of the provided
      mathematical expression.",
    "cssSelector": "#submit"
  \}
]
```
\end{alltt}
\end{llminput}
\end{attbox}
\end{document}